\pdfoutput=1
\documentclass[11pt]{article}

\usepackage{acl2023}

\usepackage{times}
\usepackage{latexsym}
\usepackage{graphicx}
\usepackage{microtype}
\usepackage{booktabs}

\usepackage[T1]{fontenc}
\usepackage[utf8]{inputenc}
\usepackage{amsmath}

\usepackage{stfloats}
\usepackage{multirow}
\usepackage{xcolor}
\newcommand*\samethanks[1][\value{footnote}]{\footnotemark[#1]}

\title{Metaphor Detection via Explicit Basic Meanings Modelling}
\author{Yucheng Li\textsuperscript{1}\thanks{\quad The two authors contributed equally to this work.}\space, Shun Wang\textsuperscript{2}\samethanks\space, Chenghua Lin\textsuperscript{2}\thanks{\quad Corresponding author}\space, Frank Guerin\textsuperscript{1}\space \\
\textsuperscript{1}~Department of Computer Science, University of Surrey, UK \\
\textsuperscript{2}~Department of Computer Science, University of Sheffield, UK \\
\texttt{\{yucheng.li, f.guerin\}@surrey.ac.uk}\\
\texttt{\{swang209, c.lin\}@sheffield.ac.uk}
}

\begin{document}
\maketitle
\begin{abstract}
% Recent computational metaphor processing uses linguistic theories such as the metaphor identification procedure (MIP) for model architecture design. 
One noticeable trend in metaphor detection is the embrace of linguistic theories such as the metaphor identification procedure (MIP) for model architecture design.
While MIP clearly defines that the metaphoricity of a lexical unit is determined based on the contrast between its \textit{contextual meaning} and its \textit{basic meaning}, existing work does not strictly follow this principle, typically using the \textit{aggregated meaning} to approximate the basic meaning of target words. In this paper, we propose a novel metaphor detection method, which models the basic meaning of the word based on literal annotation from the training set, and then compares this with the contextual meaning in a target sentence to identify metaphors. Empirical results show that our method outperforms the state-of-the-art method significantly by 1.0\% in F1 score. Moreover, our performance even reaches the theoretical upper bound on the VUA18 benchmark for targets with basic annotations, which demonstrates the importance of modelling basic meanings for metaphor detection.

%Metaphor identification is the problem of identifying metaphorical expression of words in a sentence. 
% Previous models utilize a popular linguistic theory to detect metaphor, in which metaphors can be identified by finding a contrast between a word's contextual meaning in a sentence and its original or basic meaning.

%A popular linguistic theory proposes that metaphor can be identified by finding a contrast between a word's  contextual meaning in a sentence and  its original or basic meaning. Previous models have implemented this procedure but instead of capturing the original or basic meaning, they use the most common meaning. In this paper, we propose a novel method, which models basic meaning from annotated dataset and compares it with the contextual meaning in target sentence to enhance performance on metaphor detection. Empirical results show that our method outperforms the SOTA method significantly, by 1.1 pt. Our performance even reaches a theoretical upper bound in the VUA18 benchmark for examples with complete annotation.
\end{abstract}

\section{Introduction}
Metaphors are widely used in  daily life for effective communication and vivid description. Due to their unusual and creative usage, further processes are required for machines to understand metaphors, which results in Computational Metaphor Processing (CMP), an active research direction in NLP~\cite{rai2020survey}. Recent studies demonstrate that CMP can benefit a wide range of NLP tasks including creative language generation \cite{chakrabarty2020-simile-gen,Li2022CMGenAN}, sentiment analysis \cite{li-etal-2022-secret}, and machine translation \cite{Mao2018WordEA}. Metaphor identification, aiming to detect words used metaphorically, is the very first stage in CMP.
For example, target words `\textit{attack}' or `\textit{defend}' in the context sentence \textit{``He attacks/defends her point.''}  do not literally involve \textit{physical engagement}, so they are supposed to be identified in metaphor detection for further process \cite{steen2010-from-mip-to-mipvu-most-frequent-verbal-vua}.

Linguists, philosophers and psychologists propose various ways to define metaphors, including substitution view \cite{winner1997-substitution-view}, comparison view \cite{gentner1983-comparison-view}, class inclusion view \cite{davidson1978-class-inclusion-view}, and conceptual metaphor theory \cite{lakoff2008-metaphors-we-live-by}. In contrast to these theories which are relatively complex in nature, \citet{group2007-mip} propose a simple and effective linguistic theory called Metaphor Identification Process (MIP) which can identify metaphors in unrestricted textual corpora. MIP gains increasing popularity as it detects metaphorical terms regardless of specific conceptual mapping or comparison among source and target domain, which makes the identification operational and straightforward.

According to MIP, a word is tagged as a metaphor if its contextual meaning contrast with its ``\textit{more basic meaning}''.  The basic meaning here is defined as ``\textit{more concrete; related to bodily action; more precise (as opposed to vague); historically older}'' guided by dictionaries\footnote{MIP defines basic meanings based on Macmillan and Longman Dictionary}. For example, in the sentence ``\textit{This project is such a \underline{headache}!}'', the target \textit{\underline{headache}} here is metaphorical since its contextual meaning is ``a thing or person that causes worry or trouble; a problem'', which contrasts with the more basic meaning ``a continuous pain in the head''\footnote{\url{ldoceonline.com/dictionary/headache}}.

\begin{figure*}[ht]
    \centering
    \includegraphics[width=0.9\textwidth]{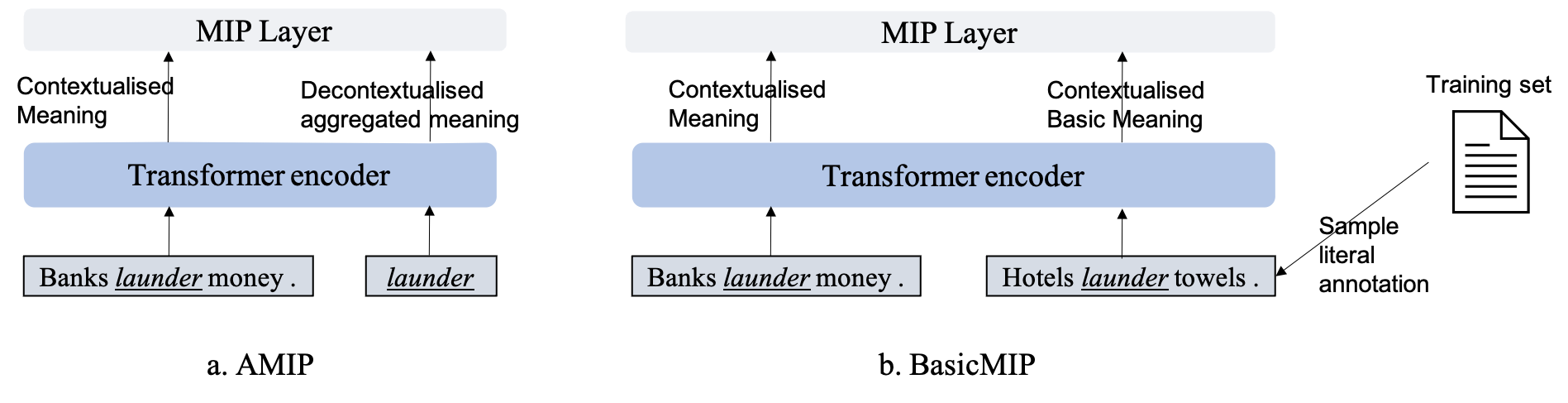}
\vspace{-1ex}    \caption{Comparison of the AMIP implementation in \cite{mao2019-mip-spv,choi2021melbert} and our BasicMIP.}
    \label{fig:BasicMIP}
\end{figure*}

Existing deep learning methods for metaphor identification usually depend on MIP in their model design~\cite{mao2019-mip-spv,choi2021melbert,song2021-mrbert,Li2023FrameBERTCM,Wang2023MetaphorDW}. However, existing works usually ignore basic meaning modelling and instead use \textit{aggregated meaning} to contrast with contextual meaning in MIP. We call the MIP in these implementations `Aggregated MIP' (AMIP).
For example, \citet{mao2019-mip-spv} and \citet{Li2023FrameBERTCM} implement MIP by contrasting contextual meaning representation with GloVe embedding and Decontextualised\footnote{which means feed the single word to pretrained language model and use the outputted vector as the representation.} RoBERTa embedding,  respectively. 
However, aggregated meaning representations, such as GloVe and decontextualised embeddings, are not the same as basic meanings in general. They usually represent a frequency-based weighted average of multiple word meanings.
In cases where the basic meaning is the most frequent, then the aggregated meaning can be a reasonable approximation to basic meaning. However, it is very common that metaphorical meanings are more frequent so that using aggregated meaning violates the fundamental rule of MIP. For example `\textit{back}' means `the rear surface of the human body' as basic meaning, but its non-basic senses, e.g. `\textit{going back}', `\textit{back up}', `\textit{back in 1960}', are more frequently used in corpora. This makes the aggregated representation of \textit{back} diverge from its basic sense, so that metaphor cannot be identified via measuring contrast with contextual meaning.

A further pitfall of previous works is that the aggregated representations used are static rather than contextualised. For example, aggregated representation GloVe and Decontextualised RoBERTa embeddings used by \citet{mao2019-mip-spv} and \citet{Li2023FrameBERTCM} are both static embedding, which are not compatible with the contextual meaning they compared to and has been shown to have worse representational quality \cite{bommasani2020-bert-decontextualised}.

In this paper, we propose a novel metaphor identification mechanism, BasicMIP, which implements MIP via direct basic meaning modelling of targets. 
BasicMIP explicitly leverages basic annotations from training set, where basic meaning of words are labeled as \texttt{literal} according to MIP theory.
First, it samples \texttt{literal} instances for each target. 
Then, the basic meaning representation of target is obtained by summing up the target embeddings of sampled \texttt{literal} instances. 
Finally, the basic representations are contrasted with their contextual meaning representation in target sentences to identify metaphors. We also present our novel metaphor detection model, BasicBERT, which not only uses BasicMIP but also inherits the AMIP module and SPV \cite[Selectional Preference Violation][]{wilks1975preferential,wilks1978making}
theory from prior works.

Extensive experiments conducted on two metaphor benchmarks show that BasicBERT significantly outperforms current SOTAs. In the VUA20 benchmark, our model exceeds MelBERT by 1\% in F1 score. In the VUA18 benchmark, our performance even reaches the theoretical upper bound for the targets with \texttt{literal} annotations in the training set. Our code and data can be found at \url{https://github.com/liyucheng09/BasicBERT}.

\section{Method}

BasicBERT model consists of three main components: 
BasicMIP, AMIP, and SPV. %, which utilise two linguistic theories for metaphor identification, namely, MIP and 
We include both AMIP and BasicMIP as some words do not have literal annotations in training set, so AMIP is an useful augmented component for these cases.
%a backup method.

\subsection{BasicMIP}
BasicMIP, as shown in Figure \ref{fig:BasicMIP}, is based on MIP, in which a target word's contextualised meaning in the current context is compared with its more basic meaning. 
% To derive the basic meaning representation of target, we first randomly sample \texttt{literal} annotation of the target from training set. Then, we consider the target within the sampled sentence is using its basic meaning and obtain contextualised basic meaning embedding via RoBERTa encoder at the end. 
\textbf{First}, the contextual meaning representation %in the current context%
is produced by feeding the current sentence to the RoBERTa network \cite{liu2019roberta}.
Formally, given a sentence $S=(w_\text{1}, ..., w_t, ..., w_n)$, where $w_t$ is the target word, we obtain representations as follows:
\begin{equation}
    H=\mathrm{RoBERTa}(\mathrm{emb}_{\text{cls}}, ..., \mathrm{emb}_{t}, ..., \mathrm{emb}_{n})
    \label{equa:roberta}
\end{equation}
Here \textsc{cls} is a special token indicating the start of an input; $\mathrm{emb}_{i}$ is the input embedding for word $w_i$; and $H=(h_{\text{cls}}, ..., h_t, ..., h_n)$ represents the output hidden states. We denote the contextual meaning embedding of $w_t$ as $v_{S, t} = h_t$ .

\textbf{Second}, to contrast the contextual meaning with the basic meaning, our model learns the basic meaning representation of the target from the training annotations. According to MIP \cite{steen2010-from-mip-to-mipvu-most-frequent-verbal-vua}, we consider targets with \texttt{literal} label to represent their basic meaning. Therefore, we sample \texttt{literal} examples of the target $w_t$ from the training set denoted as $S_b=(..., w_t, ...) \in \mathcal{U}$, where $\mathcal{U}$ is training set and $S_b$ stands for the context sentence containing a basic usage of $w_t$. Our model obtains the basic meaning embedding of $w_t$ by feeding $S_b$ to a RoBERTa encoder similar to Equation \ref{equa:roberta} and get the $t$-th output hidden state $h_t$. The final \textit{contextualised} basic representation of $w_t$ is averaged among multiple literal instances, and is formulated as $v_{B, t}$, 
which is intrinsically different to the aggregated representation of frequent meaning used in prior works.

\textbf{At last}, %we concatenate the embedding of the contextualised meaning and the basic meaning which is ready to be used in prediction layer
we compute a hidden vector $h_\text{BMIP}$ forBasicMIP, by concatenating $v_{S, t}$ and $ v_{B, t}$.
\begin{equation}
    h_{\text{BMIP}} = f_0([v_{S, t}, v_{B, t}])
    \label{equa: bmip}
\end{equation}
where $f_0(\cdot)$ denotes a linear layer to learn semantic difference between $v_{S,t}$ and $v_{B,t}$.
% Specifically, $ \mathrm{emb}_{i} = \mathrm{emb}_w + \mathrm{emb}_{pos} + \mathrm{emb}_{type}$, where $\mathrm{emb}_{w}$ is the word embedding, $\mathrm{emb}_{pos}$ the position encoding and $\mathrm{emb}_{type}$ the type encoding for  distinguishing target words within the sentence. 

\subsection{AMIP and SPV}
The AMIP implementation of MIP theory is inherited by our model, where contextual meaning and aggregated meaning of the target are compared. Here the contextual target meaning embedding of $w_t$ is $v_{S, t}$, the same as in  Equation \ref{equa: bmip}. Then, we feed the single target word $w_t$ to the RoBERTa network to derive the decontextualised vector representing the aggregated meanings of $w_t$ \cite{choi2021melbert}: $v_{F, t} = \mathrm{RoBERTa}(\mathrm{emb}_t)$.
% \begin{equation}
% \label{equa:literal target embedding}
%     V_{F, t} = \mathrm{RoBERTa\_Enc}(\mathrm{emb}_t)
% \end{equation}

The SPV theory is also employed which measures the incongruity between the contextual meaning of the target and its context. Similarly, the contextual target meaning embedding is $v_{S, t}$, and the context sentence meaning is produced by the \textsc{cls} embedding denoted as $v_S$, where $v_S = h_{cls}$. 

Finally, we compute AMIP ($h_{\text{AMIP}}$) from the contextual and aggregated target embedding, and SPV ($h_{\text{SPV}}$) from the contextual target meaning embedding and the sentence embedding. 
%This approach has been proven effective by previous works using MIP and SPV \citep{mao2019end, choi2021melbert}. 
\begin{gather}
    h_{\text{SPV}} = f_1([v_{S}, v_{S,t}]) \\
    h_{\text{AMIP}} = f_2([v_{S,t}, v_{F,t}])
\end{gather}
where $f_1{}(\cdot)$ and $f_2(\cdot)$ denote a linear layer to learn the contrast between two features.

\subsection{Prediction}
Finally, we combine three hidden vectors $h_{\text{AMIP}}$, $h_{\text{SPV}}$ and $h_{\text{BMIP}}$ to compute a prediction score $\hat{y}$, and use binary cross entropy loss to train the overall framework for metaphor prediction.
% {\small
\begin{gather}
    \hat{y}=\sigma (W^{\top }[h_{\text{BMIP}}; h_{\text{AMIP}}; h_\text{SPV}]+b) \\
    \mathcal{L}= -\sum_{i=1}^{N}[y_{i}\log\hat{y}_{i}+(1-y_{i})\log(1-\hat{y}_{i})]
\end{gather}
% }
\section{Experiments}
%\subsection{Dataset}
\noindent\textbf{Dataset.}~~We conduct experiments on two public bench datasets: 
\textbf{VUA18} \citep{leong2018report} and \textbf{VUA20} \citep{leong2020report},
%VU Amsterdam Metaphor Corpus (VUA) has been released in figurative language workshop of ACL in 2018 and 2020, which are the most popular metaphor detection benchmarks. 
which are the most popular metaphor detection benchmarks, released in the figurative language workshops of ACL in 2018 and 2020. VUA20 is an extended version of VUA18 which contains more annotations.

\vspace{1mm}

\noindent\textbf{Baselines.}~~
\textbf{RNN\_ELMo} \citep{gao2018neural} combined ELMo and BiLSTM as a backbone model.
\textbf{RNN\_MHCA} \citep{mao2019-mip-spv} introduced MIP and SPV into RNN\_ELMo and capture the contextual feature within window size by multi-head attention.
\textbf{RoBERTa\_SEQ} \citep{leong2020report} is a fine-tuned RoBERTa model in the sequence labeling setting for metaphor detection.
\textbf{MelBERT} \citep{choi2021melbert} realize MIP and SPV theories via a RoBERTa based model. 
\textbf{MrBERT} \citep{song2021-mrbert} is the SOTA on verb metaphor detection based on BERT with  verb relation encoded. 
\textbf{FrameBERT} \cite{Li2023FrameBERTCM} uses frame classes from FrameNet in metaphor detection and achieves SOTA performance on both VUA18 and VUA20.

\vspace{1mm}

\noindent\textbf{Implementation details.}~~ For target words which have no \texttt{literal} annotations in the training set, we return the decontextualised target representation as the basic meaning vector in the BasicMIP module. Therefore, the BasicMIP, in this situation, will degenerate to the AMIP implementation.

\begin{table}[tb]
\resizebox{\columnwidth}{!}{
\centering
\begin{tabular}{lcccccc}
\toprule 
 \multirow{2}{*}{Models}    &  & VUA18  &  &  & VUA20  &    \\ \cmidrule{2-7}
 
           & Prec & Rec  & F1 & Prec & Rec  & F1      \\ \midrule
RNN\_ELMo             & 71.6 & 73.6 & 72.6 & - & - & -          \\
RNN\_MHCA & 73.0 & 75.7 & 74.3 & - & -& - \\
RoBERTa\_SEQ          & 80.1 & 74.4 & 77.1 & 75.1 & 67.1 & 70.9         \\
MrBERT                  & 82.7 & 72.5 & 77.2 & - & - & - \\
MelBERT     & 80.1 & 76.9 & 78.5  & 75.9 & 69.0 & 72.3 \\
FrameBERT   & 82.7 & 75.3 & 78.8 & 79.1 & 67.7 & 73.0 \\\midrule
BasicBERT       & 79.5    & 78.5 & \textbf{79.0*} & 73.3 & 73.2 & \textbf{73.3*}         \\
w/o BasicMIP & 81.7 & 75.1 & 78.3 & 74.8 & 69.8 & 72.2 \\
% w/o AMIP & \\
\bottomrule
\end{tabular}
}
\caption{Performance comparison on VUA datasets (best results in \textbf{bold}). NB: * denotes our model outperforms the competing model with $p < 0.05$ for a two-tailed t-test.}  
\label{tabel:VUA_result}
\end{table}

\section{Results and Analysis}
\noindent\textbf{Overall results.}~~
Table~\ref{tabel:VUA_result} shows a comparison of the performance of our model against the baseline models on VUA18 and VUA20. BasicBERT outperforms all baselines on both VUA18 and VUA20, including the SOTA model MelBERT by 0.5\% and 1.0\% in F1 score, respectively. A two-tailed $t$-test was conducted based on 10 paired results (with different random seeds) between BasicBERT and the strongest baseline MelBERT on both VUA18 ($p=0.022$) and VUA20 ($p=0.006$). 

%\vspace{1mm}
\noindent\textbf{Ablation test.}~~
We also perform an ablation experiment to test the benefit of the basic modelling. As shown in Table \ref{tabel:VUA_result}, the performance of BasicBERT drops substantially when removing basic meaning modelling (w/o BasicMIP) by 0.7\% on VUA18 and 1.1\% on VUA20, respectively.

\vspace{1mm}

\begin{table}[tb]
\resizebox{\columnwidth}{!}{
\begin{tabular}{l|llcccc}
\toprule
      &  Models & Annotation   & \#sample & \#target & F1   & Acc  \\ \midrule
\multirow{4}{*}{\rotatebox{90}{VUA20}} & \multirow{2}{*}{w/  BMIP} & has literal    & 18060        & 4076        & \textbf{74.7} & 91.2 \\
&  & no literal & 4136         & 2539        & 68.2 & 86.9 \\ 
\cmidrule{2-7}
% \cmidrule{2-6}
% & BasicBERT \\
& \multirow{2}{*}{w/o BMIP} & has literal  & 18060 & 4076 & 73.3 & 91.0 \\
& & no literal & 4136 & 2539 & 68.2 & 87.6 \\ 
\midrule
\multirow{4}{*}{\rotatebox{90}{VUA18}} & \multirow{2}{*}{w/ BMIP} & has literal    & 38825        & 3874        & \textbf{81.1} & 94.7 \\
& & no literal & 5122         & 2915        & 67.3 & 87.4 \\ \cmidrule{2-7}
& \multirow{2}{*}{w/o BMIP} & has literal & 38825 & 3874   & 80.7 & 94.8 \\ 
& & no literal & 5122 & 2915 & 66.5 & 88.0 \\  \bottomrule
\end{tabular}
}
\caption{Breakdown results of BasicMIP. \textbf{has literal} indicates targets have \texttt{literal} annotations in the training set, and \textbf{no literal} indicates they have not.}
\label{tabel:breakdown}
\end{table}

\begin{table}[tb]
\centering\small
\begin{tabular}{lcc}
\toprule
Modules  & Metaphor & Literal   \\ \midrule
Contextual vs. Frequent     & 0.516    & 0.642      \\
Contextual vs. Basic & \textbf{-0.082}   & \textbf{0.809}      \\ \bottomrule
\end{tabular}
\caption{Contrast of features within AMIP and BasicMIP. The experiment is conducted on VUA20.}  
\label{tabel:cosine_similarity}
\end{table}

\noindent\textbf{Target with and without basic annotation}~~
Some target words in the test set might not have \texttt{literal} annotations in the training set. To better understand the mechanism of basic meaning modelling, we test the performance of BasicBERT on targets \textit{has} and \textit{has not} basic meaning annotations in the training data. 
%with and without basic meaning occurring in the training data. 
As shown in Table \ref{tabel:breakdown}, there are 13\% of samples in the VUA18 test set for which we cannot find a corresponding basic meaning annotation from training set. This number increases to 22\% for VUA20. We find BasicBERT gains significant improvement on targets with \texttt{literal} annotations from VUA20 via basic meaning modelling by 1.4\% in F1 score. For these targets with \texttt{literal} annotations in the VUA18 benchmark, BasicBERT gives 81.1\% in F1 score, which reaches the theoretical upper bound since the Inter-annotator agreement (IAA) value of VUA18 is around 0.8 \cite{leong2018report} (which means further improvement might lead to overfitting).

\noindent\textbf{Contrast measuring.}~~
To better compare our BasicMIP with AMIP, we conduct an experiment to directly measure the contrast between features within BasicMIP and AMIP, i.e., the contrast between the contextual and the basic meaning for BasicMIP, and the contrast between the contextual and the most frequent meaning for AMIP. Intuitively, we expect the contrast to be obvious for metaphor cases and to be slight for literal cases. Cosine distance is used to compute the contrast between two features. 
% , where $S_\text{BasicMIP} = cos(v_{S,t}, v_{B,t})$, and $S_\text{AMIP} = cos(v_{S,t}, v_{F,t})$
The contrast will fall into $(-1,1)$, smaller numbers meaning more contrasting,  larger numbers meaning less contrasting.

The results (see Table~\ref{tabel:cosine_similarity}) show that the contrast of BasicMIP features is much more obvious for metaphorical samples, and there is less contrast for literal samples compared with AMIP. Moreover, AMIP only shows a minor gap of 0.13 contrast between metaphor and literal cases. However, a significant gap of 0.89  is captured by BasicMIP between metaphor and literal cases, which demonstrates that BasicMIP learns the difference between metaphorical and literal expressions well. In summary, the results show the effectiveness of basic meaning modelling in metaphor detection.
% Therefore, the experiment confirmed that ourBasicMIP is able to learning more difference between metaphor and literal.

\noindent\textbf{Case study.}~~We perform an exploratory analysis on metaphors where BasicMIP succeeds to detect but fails without it. Prior methods might find very simple targets difficult to classify, such as \textit{\underline{see}, \underline{back}, \underline{hot}}. This is mainly because their metaphorical meanings are more frequent than their basic meanings, which leads the aggregated representations dominate by metaphorical semantics. For example, \textit{see} means \textit{look} basically. But, \textit{I see why you are angry} and \textit{this place has seen the war} are even more frequent in language corpus. Therefore, the contrast with contextual meaning tends not to indicate metaphors anymore. On the contrary, basic meaning modelling learns their basic representation by focusing literal annotations directly, which enables BasicMIP to tackle them with high accuracy (see Appendix \ref{sec:sense} for examples).
% The target word \textit{\underline{hot}} is usually difficult for previous methods but can be solved by our method with high accuracy. 
% By investigating a popular sense disambiguation corpus, Semcor \cite{miller1994-semcor}, we propose a possible reason:  \textit{hot} has several frequent meanings uniformly distributed, therefore the frequent meaning representation of \textit{get} is an average embedding. Thus the contrast with contextual meaning is less obvious by using frequent meaning representation. 
% Another type of example that is hard for prior models but can be tackled via BasicMIP is targets whose basic meaning is not the frequent meaning such as \textit{\underline{back}} (see  Appendix \ref{sec:sense} examples).
% Another kind of case is the target \textit{\underline{back}}, whose basic meaning is not the frequent meaning; this is another type of example that is hard for prior models but can be tackled viaBasicMIP (see  Appendix \ref{sec:sense} examples).

\section{Conclusion}
We proposed BasicBERT, a simple but effective approach for metaphor detection. The key feature of our method is the basic meaning modelling for metaphors from training annotations. Extensive experiments show that our model achieves best results on two benchmarks against SOTA baselines and also reaches the theoretical upper bound for instances with basic annotation. We believe our approach can be extended to other creative language with minor updates. In future, we will try apply our approach to identify other types of creative language, such as humour and sarcasm.

\section{Limitations}
This paper mainly focuses on modelling basic meaning to identify metaphors, typically learning basic meanings from literal annotations of the VUA dataset. However, our analysis reveals that the literal annotations of the VUA dataset are incomplete, which means that some words in VUA have no literal instances annotated. Although we propose using contextual word embeddings as a backup in this paper, another promising solution for this issue might be using external resources such as dictionaries. Leveraging dictionaries is commonly used to assist manual metaphor detection, so it could also help our BasicMIP mechanism to generalise. We leave this for future work.

\bibliography{anthology,custom}
\bibliographystyle{acl_natbib}

\appendix

\newpage

\section{Examples of targets \textit{get} and \textit{back}}
\label{sec:sense}
\begin{figure*}
    \centering
    \includegraphics[width = 0.6\textwidth]{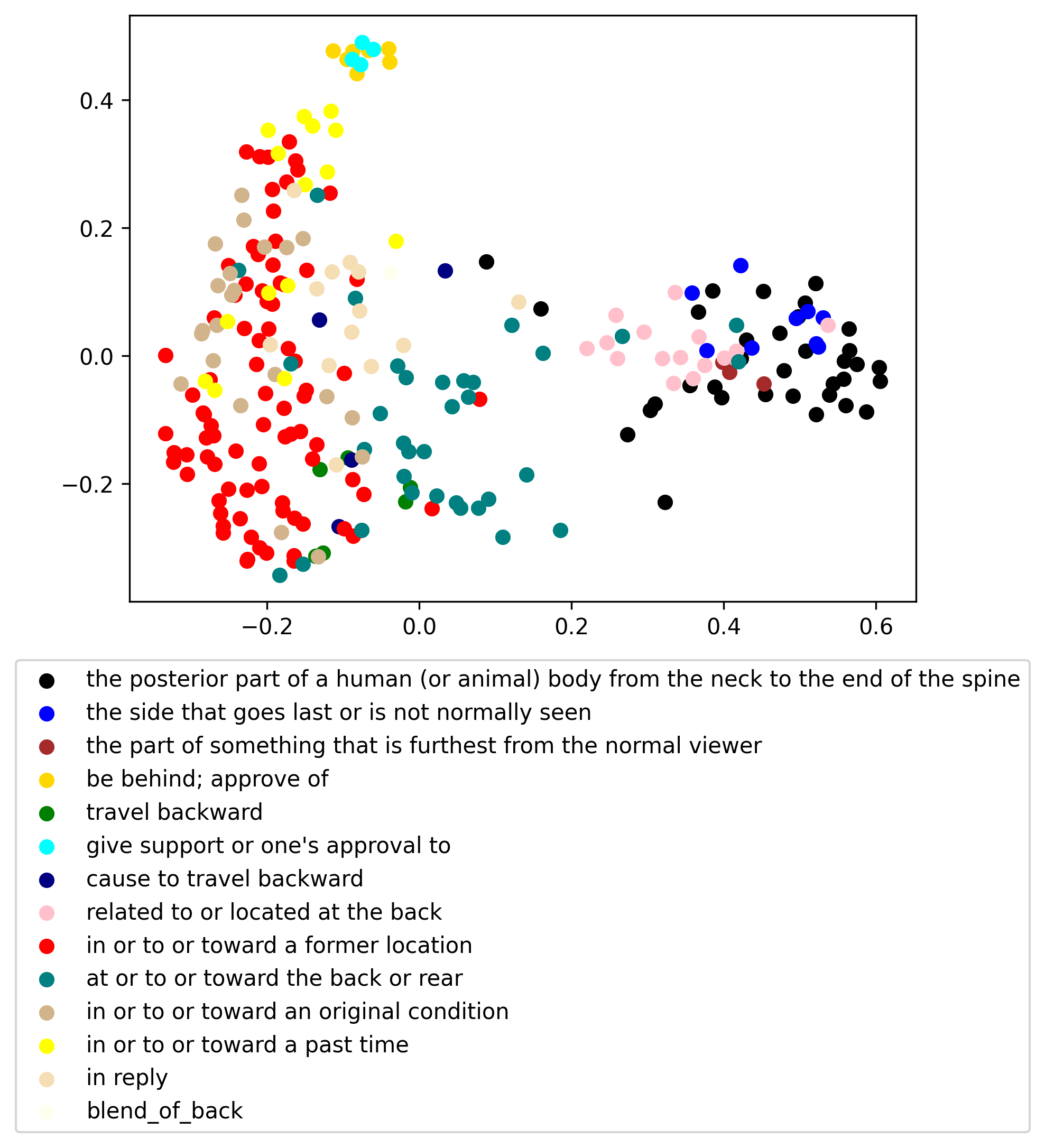}
    \caption{Senses of \textit{back} from word sense disambiguation dataset Semcor.}
    \label{fig:back}
\end{figure*}

\begin{figure*}
    \centering
    \includegraphics[width = 0.6\textwidth]{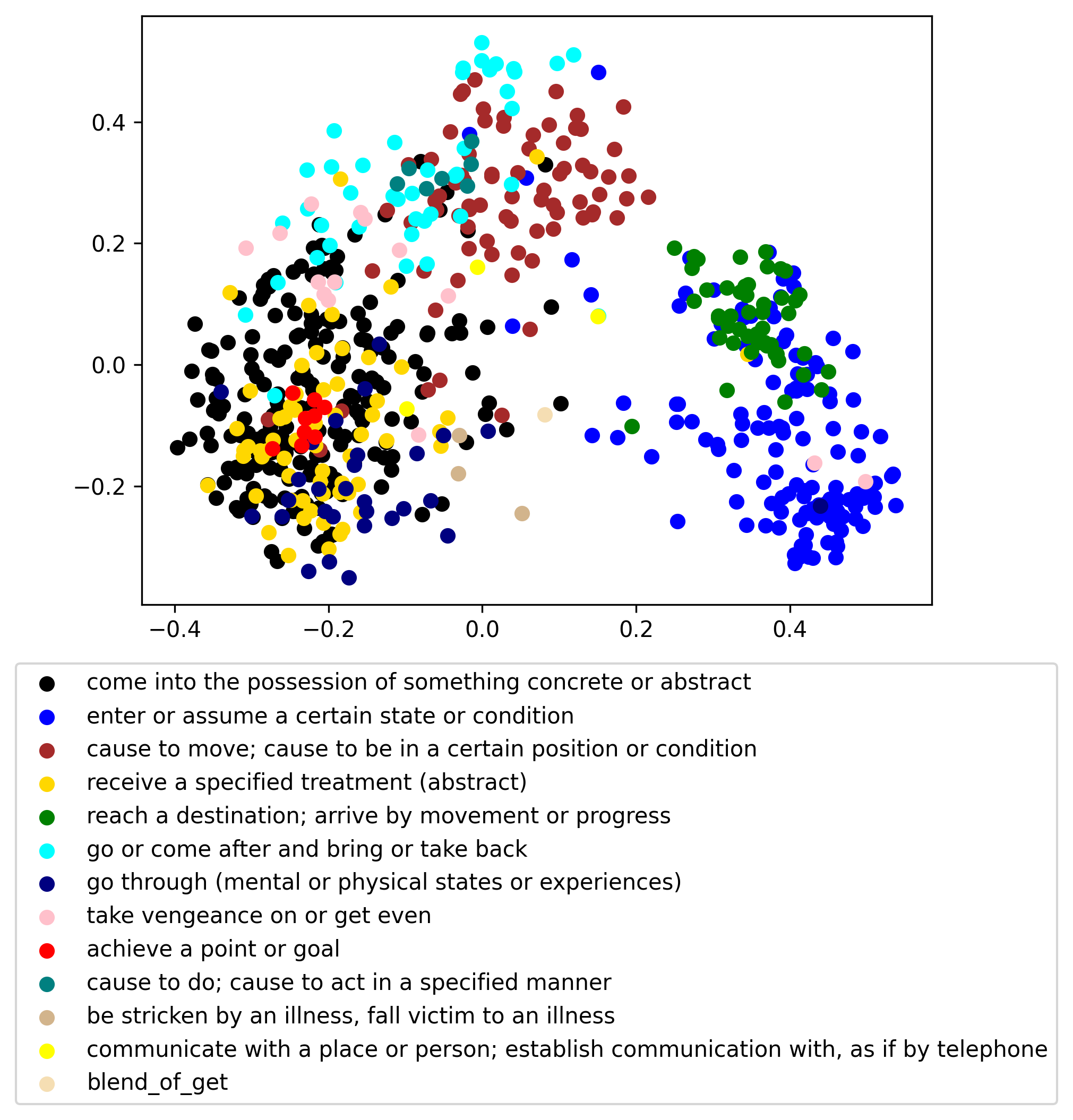}
    \caption{Senses of \textit{get} from word sense disambiguation dataset Semcor.}
    \label{fig:get}
\end{figure*}

Table \ref{tabel:cases} shows cases where previous methods fails but ours successes. Corresponding sentences with basic usage of target from training set are also included. We also show word senses illustration in Figure \ref{fig:back} and Figure \ref{fig:get}. The figure is drawn via RoBERTa embedding and PCA techniques. We can see the most frequent meaning of \textit{back} is \textit{`former location'} and \textit{`travel backward'} instead of the basic meaning \textit{`human body'}. And the meanings of \textit{get} are almost equally frequent.

\begin{table*}[tb]
\resizebox{\textwidth}{!}{
\centering
\begin{tabular}{l|p{8cm}p{8cm}}
\toprule 
 Target    & Cases  & Basic Examples  \\ \midrule
 \multirow{3}{*}{get}  & we will , i 'm just saying we do wan na \textcolor{red}{get} into cocktail & where do you \textcolor{red}{get} your carrots from ? \\
  & they 're watching neighbours come on , \textcolor{red}{get} up you lazy bugger ! & and you 'll \textcolor{red}{get} a separate room  \\ 
  &  oh we did n't \textcolor{red}{get} much further on there , what we started with this morning. & i 'm gon na get some cleaning , i 'll \textcolor{red}{get} some cleaning fluid this week . \\ \midrule
  \multirow{2}{*}{back} & why ca n't they take it through the \textcolor{red}{back} door and up the stair ? &  within 10 minutes i had turned my \textcolor{red}{back} on the corduroy battalions of trees and was striding under a still. \\
  &  they are unlikely to find a place to do so which is not in somebody 's \textcolor{red}{back} yard . & on the edge of the lawn with his \textcolor{red}{back} to the cedar tree . \\
  \bottomrule
\end{tabular}
}
\caption{Cases study of targets \textit{get} and \textit{back}}  
\label{tabel:cases}
\end{table*}

\begin{table}[tb]
%\resizebox{\columnwidth}{!}{
\centering
\begin{tabular}{l|c}
\toprule 
 Hardware & TITAN RTX  \\ \midrule
 Runtime/epoch & 50 min  \\ \midrule
 Parameters  & 252,839,426 \\
  \bottomrule
\end{tabular}
%}
\caption{Experiment details}  
\label{tabel:experiment}
\end{table}

% This is an appendix.

\end{document}